\documentclass[runningheads]{llncs}
\usepackage{graphicx,url}
\usepackage[american]{babel}
\usepackage[utf8]{inputenc}  
\usepackage[utf8]{inputenc}
\usepackage{csquotes}
\usepackage{esvect}
\usepackage{setspace}
\usepackage{amsmath,amssymb}
\usepackage{amsfonts}
\usepackage{graphicx}
\usepackage{float}
\usepackage{pdfpages}
\usepackage{hyperref}
\usepackage{indentfirst}
\usepackage{multirow}
\usepackage{cellspace}
\usepackage{graphicx}
\usepackage{xcolor}
\usepackage{amssymb}
\usepackage{amsmath}
\newcommand{\ii}{\boldsymbol{i}}
\newcommand{\jj}{\boldsymbol{j}}
\newcommand{\kk}{\boldsymbol{k}}

\newcommand{\euler}{e}
\begin{document}

\title{Extending the Universal Approximation Theorem for a Broad Class of Hypercomplex-Valued Neural Networks
\thanks{This work was supported in part by the National Council for Scientific and Technological Development (CNPq) under grant no 315820/2021-7, the S\~ao Paulo Research Foundation (FAPESP) under grant no 2022/01831-2, and the Coordena\c{c}\~ao  de Aperfei\c{c}oamento  de Pessoal de N\'ivel Superior - Brasil (CAPES) - Finance Code 001.}
}
\titlerunning{Approximation Theorem for Hypercomplex-valued Neural Networks}
%
\author{Wington L. Vital\orcidID{0000-0003-1634-4441} \and Guilherme Vieira\orcidID{0000-0003-3361-6154} \and Marcos Eduardo Valle\orcidID{0000-0003-4026-5110}}
\institute{Universidade Estadual de Campinas, Campinas, Brazil.
\newline \email{w265003@dac.unicamp.br, vieira.g@dac.unicamp.br, and valle@dac.unicamp.br}
}
\maketitle              

\begin{abstract}
The universal approximation theorem asserts that a single hidden layer neural network approximates continuous functions with any desired precision on compact sets. As an existential result, the universal approximation theorem supports the use of neural networks for various applications, including regression and classification tasks. The universal approximation theorem is not limited to real-valued neural networks but also holds for complex, quaternion, tessarines, and Clifford-valued neural networks. This paper extends the universal approximation theorem for a broad class of hypercomplex-valued neural networks. Precisely, we first introduce the concept of non-degenerate hypercomplex algebra. Complex numbers, quaternions, and tessarines are examples of non-degenerate hypercomplex algebras. Then, we state the universal approximation theorem for hypercomplex-valued neural networks defined on a non-degenerate algebra.

\end{abstract}

\section{Introduction}
Artificial neural networks are computational models created to emulate the behavior of biological neural networks. Their origins are tied back to the pioneer works of McCulloch and Pitts \cite{McCulloch1943}, and Rosenblatt \cite{rosenblatt1958perceptron}. Since then, many applications have emerged in various fields, such as computer vision, physics, control, pattern recognition, economics, and many applications in the medical field. Neural networks are known for being approximators with adjustable capability. Thus, a major interest in the topic of neural networks is that of approximating a generic class of functions with arbitrary precision. The approximation capability of neural networks was initially motivated by representation theorems and the need to provide its theoretical justification  
\cite{Arnold2009,kolmogorov1957representation}.

As far as we know, the starting point of the approximation theory for neural networks was the \textit{universal approximation theorem} formulated by Cybenko in the late 1980s \cite{Cybenko1989ApproximationFunction}. In a few words, Cybenko showed that a single hidden layer real-valued multilayer perceptron (MLP) equipped with a sigmoid activation function could approximate continuous function to any desired precision in a compact set. A few years later, Cybenko's universal approximation theorem was generalized to real-valued MLP models with any non-constant bounded activation function \cite{HORNIK1991251}. Recently, many researchers addressed the approximation capabilities of neural networks, including deep and shallow models based on piece-wise linear activation functions such as the widely used rectified linear unit $\mathtt{ReLU}$ \cite{PETERSEN2018296}.

In the 1990s, Arena et al. extended the universal approximation theorem for complex and quaternion-valued single hidden layer feedforward networks with the so-called split activation functions \cite{ARENA1997335,arena1998neural}. This significant breakthrough was vital in formulating universal approximation theorems for other hypercomplex-valued neural networks, such as the hyperbolic and tessarine-valued networks \cite{buchholz2000hyperbolic,eniac}. In particular, the universal approximation theorem has been successfully extended for neural networks defined on Clifford algebras by Buchholz and Sommer in the early 2000s \cite{Buchholz2001}. 

Despite the results mentioned above, there is a lack of a more general version of the universal approximation theorem. This work extends the universal approximation theorem to a broad class of hypercomplex algebras. Indeed, we consider a broad framework for hypercomplex numbers, which includes the most widely used algebras as particular instances \cite{Catoni2008TheSpace-Time,kantor1989hypercomplex}. Then, we address the problem of approximating a continuous hypercomplex-valued function on a compact subset by a hypercomplex-valued multilayer perceptron ($\mathbb{H}$MLP). The theoretical results present in this paper justifies some recent successfull applications of neural networks based on hypercomplex algebras beyond complex numbers and quaternions \cite{grassucci22prl,grassucci2022lightweight,Takahashi2021ComparisonControl,vieira22wcci_arxiv,vieira2022general}.

The paper is organized as follows: Section \ref{sec:review} briefly reviews concepts regarding hypercomplex algebras. Section \ref{sec:UATS_diversos} reviews the MLP architecture and the existing universal approximation theorems. The main result of this work, namely, the universal approximation theorem for a broad class of hypercomplex-valued neural networks, is given in Section \ref{sec:UAT_HVNN}. We would like to point out that we omitted the results' proofs due to the page limit. The paper finishes with some concluding remarks in Section \ref{sec:concluding}.

\section{A brief review of hypercomplex algebras} \label{sec:review}

Let us start by recalling the basic theory of hypercomplex algebras \cite{Catoni2008TheSpace-Time,kantor1989hypercomplex}. This theory is of paramount importance to the main results of this work, which will be detailed further in Section \ref{sec:UAT_HVNN}.

The hypercomplex algebras considered in this paper are defined over the field $\mathbb{R}$, but it is worth mentioning that it is possible to work with such algebras over any field. For a more general extensive approach to hypercomplex algebra concepts, please refer to \cite{Catoni2008TheSpace-Time,kantor1989hypercomplex}.

A hypercomplex number $x$ has a representation in the form
\begin{eqnarray}\label{eq:hyper_number}
x=x_{0}+x_{1} \boldsymbol{i}_{1}+\ldots+x_{n} \boldsymbol{i}_{n},
\end{eqnarray}
where $x_{0}, x_{1}, \ldots, x_{n} \in \mathbb{R}$. The elements $\boldsymbol{i}_{1}, \boldsymbol{i}_{2}, \ldots, \boldsymbol{i}_{n}$ are called hyperimaginary units. 

The addition of hypercomplex numbers is done component by component, that is,
\begin{equation}\label{eq2}
x+y=\left(x_{0}+y_{0}\right)+\left(x_{1}+y_{1}\right) \boldsymbol{i}_{1}+\ldots+\left(x_{n}+y_{n}\right) \boldsymbol{i}_{n},
\end{equation}
for hypercomplex numbers $x=x_{0}+x_{1} \boldsymbol{i}_{1}+\ldots+x_{n} \boldsymbol{i}_{n}$ and $y=y_{0}+y_{1} \boldsymbol{i}_{1}+\ldots+y_{n} \boldsymbol{i}_{n}$.

The multiplication of two hypercomplex numbers is performed distributively using the product of the hyperimaginary units. Precisely, the product of two hypercomplex units is defined by 
\begin{eqnarray}\label{eq:im_unit_prod}
\boldsymbol{i}_{\alpha}\boldsymbol{i}_{\beta} := p_{\alpha \beta,0} + p_{\alpha \beta,1}\boldsymbol{i}_{1} + \ldots + p_{\alpha \beta,n}\boldsymbol{i}_{n}. 
\end{eqnarray}
for all $\alpha,\beta =1, \ldots, n$ and $p_{\alpha \beta, \gamma} \in \mathbb{R}$ with $\gamma = 0,1, \ldots, n$ . In this way, the multiplication of the hypercomplex numbers $x=x_{0}+x_{1} \boldsymbol{i}_{1}+\ldots+x_{n} \boldsymbol{i}_{n}$ and $y=y_{0}+y_{1} \boldsymbol{i}_{1}+\ldots+y_{n} \boldsymbol{i}_{n}$ is computed as follows
\begin{align}\label{eq:prod}
xy &= \left( x_0y_0 + \sum_{\alpha,\beta = 1}^n x_\alpha y_{\beta} p_{\alpha\beta,0}\right) \nonumber \\  & + \left(x_0y_1+x_1y_0 + \sum_{\alpha,\beta=1}^n x_\alpha y_\beta p_{\alpha\beta,1} \right) \boldsymbol{i}_1 + \ldots  \nonumber \\
& + \left(x_0y_n+x_ny_0 + \sum_{\alpha,\beta=1}^n x_\alpha y_\beta p_{\alpha\beta,n} \right) \boldsymbol{i}_{n}. 
\end{align}

A hypercomplex algebra, which we will denote by $\mathbb{H}$, is a hypercomplex number system equipped with the addition \eqref{eq2} and the multiplication \eqref{eq:prod}. 

We would like to remark that the product of a hypercomplex number $x = x_{0}+ x_{1} \boldsymbol{i}_{1}+\cdots+ x_{n} \boldsymbol{i}_{n}$ by a scalar $\alpha \in \mathbb{R}$, given by  
\begin{equation} \label{eq:scalar_product}
\alpha x=\alpha x_{0}+\alpha x_{1} \boldsymbol{i}_{1}+\cdots+\alpha x_{n} \boldsymbol{i}_{n},
\end{equation}
can be derived from \eqref{eq:prod} by identifying $\alpha \in \mathbb{R}$ with the hypercomplex number $\alpha + 0\boldsymbol{i}_1+ \cdots+ 0\boldsymbol{i}_n \in \mathbb{H}$. As a consequence, a hypercomplex algebra $\mathbb{H}$ is a vector space with the addition and scalar product given by \eqref{eq2} and \eqref{eq:scalar_product}. Moreover, $\tau = \{1,\boldsymbol{i}_1,\ldots,\boldsymbol{i}_n\}$ is the canonical basis for $\mathbb{H}$. The canonical basis $\tau$ yields a one-to-one correspondence between a hypercomplex number $x = x_0 + x_1 \boldsymbol{i}_1 + \cdots + x_n \boldsymbol{i}_n$ and a vector $[x]_\tau = (x_0,x_1,\ldots,x_n) \in \mathbb{R}^{n+1}$. Using the such correspondence, we define the absolute value $|x|$ of a hypercomplex number $x \in \mathbb{H}$ as the Euclidean norm of $[x]_\tau$, that is,
\begin{equation} \label{eq:absolute_value}
|x| := \|[x]_\tau\|_2 = \sqrt{x_0^2+x_1^2 + \cdots + x_n^2}.
\end{equation}
Concluding, there exists an isomorphism between $\mathbb{H}$ and $\mathbb{R}^{n+1}$. However, beyond its vector space structure, an hypercomplex algebra $\mathbb{H}$ is equipped with a multiplication given by \eqref{eq:prod}.  

Complex numbers ($\mathbb{C}$), quaternions ($\mathbb{Q}$), and octonions ($\mathbb{O}$) are examples of hypercomplex algebras. Hyperbolic numbers ($\mathbb{U}$), dual numbers ($\mathbb{D}$), and tessarines ($\mathbb{T}$) are also hypercomplex algebras. The following examples illustrate further some hypercomplex algebras.

\begin{example}
Complex, hyperbolic, and dual numbers are hypercomplex algebras of dimension 2, i.e., the elements of these algebras are of the form $x = x_{0} + \ii x_{1}$. They differ in the value of $\ii^2$. The most well-known of these 2-dimensional (2D) hypercomplex algebras is the complex numbers where $\ii^2 = -1$. Complex numbers play a key role in physics, electromagnetism, and electrical and electronic circuits. In contrast, hyperbolic numbers have $\ii^2 = 1$ and have important connections with abstract algebra, ring theory, and special relativity \cite{Catoni2008TheSpace-Time}. Lastly, dual numbers are a degenerate algebra in which $\ii^2 = 0$. 
\end{example}

\begin{example}
Quaternions are a 4D hypercomplex algebra denoted by $\mathbb{Q}$. The quaternion elements are $x = x_0 + x_1 \ii + x_2 \jj + x_3 \kk$, where $\ii \equiv \boldsymbol{i}_1$, $\jj \equiv \boldsymbol{i}_2$, $\kk \equiv \boldsymbol{i}_3$ are the hyperimaginary units. The quaternion product is associative and anticommutative, and is of particular interest to describe rotations in the 3D Euclidean space $\mathbb{R}^3$. Formally, we have:
\begin{equation}
    \ii^2 = \jj^2 = \kk^2 = -1, \quad \ii \jj = \kk, \quad \mbox{and} \quad \jj\ii = -\kk.
\end{equation}
Together with complex numbers, quaternion is one of the most well-known hypercomplex algebras. Quaternions has seen applications in many fields ranging from physics to computer vision and control due to the intrinsic relation between movement in the 3D space and quaternion product.
\end{example}

\begin{example}
Cayley–Dickson algebras are a family of hypercomplex algebras that contains the previously mentioned complex and quaternions as particular instances. The Cayley-Dickson algebras are produced by an iterative parametric process \cite{albert42} that generates algebras of doubling dimension, i.e., these algebras always have a dimension equal to a power of $2$. Cayley-Dickson algebras have been successfully used to implement efficient neural network models for color image processing tasks \cite{vieira2022general}.
\end{example}

\begin{example}
The tessarines $\mathbb{T}$ are a commutative 4D algebra similar to the quaternions, hence they are often referred to as \textit{commutative quaternions} \cite{CERRONI2017232}. The tessarines elements are $x = x_0 + x_1 \ii + x_2 \jj + x_3 \kk$, where $\ii \equiv \boldsymbol{i}_1$, $\jj \equiv \boldsymbol{i}_2$, $\kk \equiv \boldsymbol{i}_3$ are the hyperimaginary units. Unlike the quaternions, we have:
\begin{equation}
    \ii^2 = -1, \; \jj^2 = 1, \; \kk^2 = -1, \quad \mbox{and} \quad \ii \jj = \jj \ii = \kk.
\end{equation}
Like the quaternions, tessarines have been used for digital signal processing \cite{1306653,alfsmann2006families}. A recent paper by Senna and Valle addressed tessarine-valued deep neural networks, which outperformed real-valued deep neural networks for image processing and analysis tasks \cite{Eniacc}.
\end{example}

\begin{example}
The Klein four-group $\mathbb{K}_{4}$ is a 4D hypercomplex algebra whose imaginary unit are self-inverse, i.e. $\boldsymbol{i}^{2}= \boldsymbol{j}^{2}= \boldsymbol{k}^{2}=1$ and $\ii \jj = \kk$. Besides the theoretical studies in symmetric group theory \cite{huang2013klein}, the Klein four-group has been used for the design of hypercomplex-valued Hopfield neural networks \cite{KOBAYASHI2020123}.  
\end{example}

\begin{example}
Besides quaternions, tessarines, and the Klein four-group, the hyperbolic quaternions are a 4D non-associative and anticommutative hypercomplex algebra whose hypercomplex units satisfy
\begin{equation}
    \ii^2 = \jj^2 = \kk^2 = 1, \;
    \ii \jj = \kk = -\jj \ii,  \;
    \jj \kk = \ii = -\kk \jj, \;\mbox{ and }\;
    \kk \ii = \jj = -\ii \kk.
\end{equation}
Among others 4D hypercomplex algebras, the hyperbolic quaternions have been used to design a servo-level robot manipulator controller by Takahashi \cite{Takahashi2021ComparisonControl}.  
\end{example}

\begin{example}
Clifford algebras are an important family of hypercomplex algebras with interesting geometric properties and a wide range of applications \cite{Breuils2022NewAlgebra,hitzer13}. A Clifford algebra is generated from the vector space $\mathbb{R}^n$ equipped with a quadratic form $Q:\mathbb{R}^n \to \mathbb{R}$ \cite{Buchholz2008OnPerceptrons,vaz16}. Precisely, the Clifford algebra $C\ell_{p,q,r}$, where $p$, $q$, and $r$ are non-negative integers such that $p+q+r=n$,
is constructed from an orthonormal basis $\{e_1,e_2,\ldots,e_n\}$ of $\mathbb{R}^n$ such that 
\begin{equation} Q(e_i+e_j)=Q(e_i)+Q(e_j) \quad \mbox{and} \quad 
    Q(e_i) = \begin{cases}
    +1, & 1 \leq i \leq p,\\
    -1, & p+1 \leq i \leq q,\\
    0, & p+q+1 \leq i \leq n.
    \end{cases}
\end{equation}
In particular, the Clifford algebra $C\ell_{0,1,0}$ is equivalent to the complex numbers, $C\ell_{1,0,0}$ is equivalent to the hyperbolic numbers, and $C\ell_{0,2,0}$ is equivalent to the quaternions. A Clifford algebra is degenerate if $r>0$. A non-degenerate Clifford algebra $C\ell_{p,q,0}$ is also denoted by $C\ell_{p,q}$, that is, $C\ell_{p,q} \equiv C\ell_{p,q,0}$.
\end{example}

The examples above present a handful of algebras with different sets of properties or lack thereof. While complex, hyperbolic, dual numbers, tessarines and the Klein group are commutative, the quaternions and general Clifford algebras are not. The hyperbolic quaternions and the octonions, a well-known Cayley-Dickson 8D hypercomplex algebra, are not associative. The hyperbolic numbers present non-null zero divisors. Only a few properties are observed across all hypercomplex number systems $\mathbb{H}$. Notably, the identity $(\omega x)(\eta y) = (\omega \eta)(xy)$ holds for all $x,y \in \mathbb{H}$ and $\omega, \eta \in \mathbb{R}$. Also, we have distributivity as $x(y+w) = xy + xw$ and $(y+w)x = yx+ wx$, for all $x,y,z \in \mathbb{H}$.

\section{Some Approximation Theorems from the Literature}\label{sec:UATS_diversos}


A multilayer perceptron (MLP) is a feedforward artificial neural network architecture with neurons arranged in layers. Each neuron in a layer is connected to all neurons in the previous layer, hence this model is also known as fully-connected or dense. The feedforward step through a MLP with a single hidden-layer with $M$ neurons can be described by a finite linear combination of the hidden neurons outputs. Formally, the output of a single hidden-layer MLP network $\mathcal{N}_\mathbb{R}(\boldsymbol{x})$ is given by
\begin{equation}\label{eq:mlp_real}
\mathcal{N}_\mathbb{R}(\boldsymbol{x}) = \sum_{i=1}^{M} \alpha_{i} \phi(\boldsymbol{y}_{i}^{T} \cdot \boldsymbol{x} + \theta_{i}),
\end{equation}
where $\boldsymbol{x} \in \mathbb{R}^{N}$ represents the input to the neural network, $\boldsymbol{y}_{i} \in \mathbb{R}^{N}$ and $\alpha_{i} \in \mathbb{R}$ are the weights between input and hidden layers, and hidden and output layers, respectively. Moreover, $\theta_{i} \in \mathbb{R}$ is the bias terms for the $i$th neuron in the hidden layer and $\phi: \mathbb{R} \to \mathbb{R}$ is the activation function.

The class of all functions that can be obtained using a MLP with activation function $\phi$ will be denoted by
\begin{equation}\label{eq:class_MLP_real}
    \mathcal{H}_\mathbb{\phi} = \left\{ \mathcal{N}_\mathbb{R}(\boldsymbol{x}) = \sum_{i=1}^{M} \alpha_{i} \phi(\boldsymbol{y}_{i}^{T} \cdot \boldsymbol{x} + \theta_{i}): M \in \mathbb{N}, \boldsymbol{y}_{i} \in \mathbb{R}^N, \alpha_{i}, \theta_{i} \in \mathbb{R} \right\}.
\end{equation}

Sigmoid functions are widely used activation functions and include the logistc function defined by  
\begin{equation}
\label{eq:sigmoid}
    \sigma(x)=\frac{1}{1+\euler^{-x}}, \quad \forall x \in \mathbb{R},
\end{equation}
as a particular instance. Besides sigmoid functions, modern neural networks also use the rectified linear unit $\mathtt{ReLU}$ as activation function, which is defined as follows for all $x \in \mathbb{R}$:
\begin{equation} 
\mathtt{ReLU}(x)= \begin{cases}x, & \text { if } x>0, \\ 0, & \text { if } x \leq 0.\end{cases} 
\end{equation}
The key interest in the usage of activation functions is to discriminate inputs. We review this key property below, in which we denote by $\mathcal{C}(K)$ the class of all continuous functions on a compact subset $K \subset \mathbb{R}^N$.



\begin{definition}[Discriminatory Function] \label{definição1}
Consider a real-valued function $\phi: \mathbb{R} \rightarrow \mathbb{R}$ and let $K \subset \mathbb{R}^{N}$ be a compact. The function $\phi$ is said to be discriminatory if, for a finite signed regular Borel measure $\mu$ on $K$, the following holds
\begin{eqnarray}
 \int_{K} \phi(\boldsymbol{y}^{T} \cdot \boldsymbol{x} + \theta) d\mu(\boldsymbol{x}) = 0,  \quad \forall \boldsymbol{y} \in \mathbb{R}^{N} \; \text{and} \; \forall \theta \in \mathbb{R},
\end{eqnarray}
if, and only if, $\mu$ is the zero measure, i.e., $\mu = 0$.
\end{definition}
The sigmoid and $\mathtt{ReLU}$ functions defined above are examples of discriminatory activation functions \cite{Cybenko1989ApproximationFunction,guilhoto2018overview}. More generally, Hornik showed that bounded non-constant real-valued functions are discriminatory \cite{HORNIK1991251}.



The next theorem, published in 1989, establishes the universal approximation property for real-valued networks. Note that Definition \ref{definição1} plays a key role in establishing the result proved by Cybenko \cite{Cybenko1989ApproximationFunction}.

\begin{theorem}[Universal Approximation Theorem \cite{Cybenko1989ApproximationFunction}]\label{teo:TAU_real}
Consider a compact $K \subset \mathbb{R}^N$ and let $\phi:\mathbb{R} \to \mathbb{R}$ be a continuous discriminatory function. The class of all real-valued neural networks defined by \eqref{eq:class_MLP_real} 
is dense in $\mathcal{C}(K)$, the set of all real-valued continuous functions on $K$. In other words, given a real-valued continuous-function $f_\mathbb{R}:K \to \mathbb{R}$ and $\epsilon>0$, there is  
a single hidden-layer MLP network given by \eqref{eq:mlp_real}
such that
\begin{equation}
    | f_\mathbb{R}(\boldsymbol{x})-\mathcal{N}_\mathbb{R}(\boldsymbol{x})|< \epsilon, \quad \forall \boldsymbol{x} \in K.
\end{equation}
\end{theorem}


Over the following decades, the universal approximation property was proven for neural networks with values in several other algebras. We highlight some of these works in the remainder of this section.

\subsection{Complex-valued case}\label{subsec:complex}

The structure of a complex-valued MLP ($\mathbb{C}$MLP) is equivalent to that of a real-valued MLP, except that input and output signals, weights and bias are complex numbers instead of real values. Additionally, the activation functions are complex-valued functions \cite{arena1998neural}. Note that the logistic function given by \eqref{eq:sigmoid} can be generalized to complex parameters using Euler's formula  $ \euler^{x\ii} = \cos (x) + \ii \sin (x)$ as follows for all $x \in \mathbb{C}$:
\begin{eqnarray} \label{eq:sigmoid_C}
\sigma(x)=\frac{1}{1+\euler^{-x}}.
\end{eqnarray}

However, in 1998, Arena \textit{et al.} noted that the universal approximation property in the context of the $\mathbb{C}$MLP network with the activation function \eqref{eq:sigmoid_C} is generally not valid \cite{arena1998neural}. Nonetheless, they proved that the split activation function 
\begin{eqnarray}\label{eq:split_sigmoid}
\sigma(x) = \frac{1}{1+ \euler^{-x_{0}}}+\ii \frac{1}{1+ \euler^{-x_{1}}}
\end{eqnarray} 
for $x=x_{0}+ \ii x_{1} \in \mathbb{C}$ is discriminatory. Moreover, they generalized Theorem \ref{teo:TAU_real} for $\mathbb{C}$MLP networks with split sigmoid activation functions \cite{arena1998neural}.

\subsection{Quaternion-valued case}\label{subsec:quat}

In the same vein, Arena \textit{et al.} also defined quaternion-valued MLP ($\mathbb{Q}$MLP) by replacing the real input and output, weights and biases, by quaternion numbers. They then proceeded to prove that $\mathbb{Q}$MLPs with a single hidden layer and split sigmoid activation function
\begin{equation}
    \sigma(x) =  \frac{1}{1+ \euler^{-x_{0}}}+\ii \frac{1}{1+ \euler^{-x_{1}}} + \jj  \frac{1}{1+ \euler^{-x_{2}}}+\kk \frac{1}{1+ \euler^{-x_{3}}},
\end{equation}
for $x = x_0 + \ii x_1 + \jj x_2 + \kk x_3 \in \mathbb{Q}$, are universal approximators in the set of continuous quaternion-valued functions \cite{ARENA1997335}.

\subsection{Hyperbolic-valued case}\label{subsec:hyperbolic}

In the year 2000, Buchholz and Sommer introduced a MLP based on hyperbolic numbers, the aptly named hyperbolic multilayer perceptron ($\mathbb{U}$MLP). This network equipped with a split logistic activation function given by  \eqref{eq:split_sigmoid} is also a universal approximator \cite{buchholz2000hyperbolic}. Buchholz and Sommer provided experiments highlighting that the $\mathbb{U}$MLP can learn tasks with underlying hyperbolic properties much more accurately and efficiently than $\mathbb{C}$MLP and real-valued MLP networks.

\subsection{Tessarine-valued case}\label{subsec:tessarine}

Recently, Carniello \textit{et al.} experimented with networks with inputs, outputs and parameters in the tessarine algebra \cite{eniac}. The researchers proposed the $\mathbb{T}$MLP, a MLP architecture similar to the complex, quaternion and hyperbolic MLPs mentioned above but based on tesarines. The authors then proceeded to show that the proposed $\mathbb{T}$MLP is a universal approximator for continuous functions defined on compact subsets of $\mathbb{T}$ with sigmoid and the $\mathtt{ReLU}$ activation functions. Experiments show that the tessarine-valued network is a powerful approximator, presenting superior performance when compared to the real-valued MLP in a task of approximating tessarine functions \cite{eniac}.

\subsection{Clifford-valued case}\label{subsec:clifford}

In 2001, Buchholz and Sommer worked with a class of neural networks based on Clifford algebras \cite{Buchholz2001}. They found that the universal approximation property holds for MLPs based on non-degenerate Clifford algebra. In addition they pointed out that degenerate Clifford algebras may lead to models without universal approximation capability.

It is worth noting that Buchholz and Sommer considered sigmoid activation functions. However, it is possible to show that the split $\mathtt{ReLU}$ activation function is discriminatory in a Clifford algebra. Hence, Clifford MLPs are universal approximators with the the split $\mathtt{ReLU}$ activation function as well.

\section{Universal Approximation Theorem for Hypercomplex-Valued Neural Networks} \label{sec:UAT_HVNN}

This section deals with the extension of the universal approximation theorem to a wide class of artificial neural networks with hypercomplex values. This is the main result of this work, which is based on the concept of non-degenerate hypercomplex algebra.


\subsection{Non-degenerate Hypercomplex Algebras}

Let us start by introducing preliminary results and some core definitions that lead us to the main result. This subsection relies on the hypercomplex algebra concepts detailed in Section \ref{sec:review} and linear algebra \cite{hoffman1971linear}.



A linear operator on a hypercomplex algebra $\mathbb{H}$ is an operator $T: \mathbb{H} \to \mathbb{H}$ such that $T(\alpha x + y) = \alpha T (x) + T(y)$ for all $x , y\in \mathbb{H}$ and $\alpha \in \mathbb{R}$ \cite{kantor1989hypercomplex}.

A bilinear form on $\mathbb{H}$ is a mapping $B: \mathbb{H} \times \mathbb{H} \to \mathbb{R}$ such that
\begin{equation}
    B(c_{1}x_{1} + c_{2}x_{2}, v) = c_{1}B(x_{1},v) + c_{2}B(x_{2} ,v),
\end{equation}
and
\begin{equation}
    B(x,d_{1}v_{1} +d_{2}v_{2}) = d_{1}B(x,v_{1}) + d_{2}B(x,v_{ 2}),
\end{equation}
hold true for any $x_{1}, x_{2}, v_{1}, v_{2} , x, v \in \mathbb{H}$ and $c_{1}, c_{2}, d_{1}, d_{2} \in \mathbb{R}$. In words, a bilinear form is linear in both its arguments.

The following preliminary result consists of a theorem linking the hypercomplex algebra product given by \eqref{eq:prod} to bilinear forms. This result also leads to matrix representations of \eqref{eq:prod}.


\begin{theorem}\label{thm:prod_bilin}
Let $\mathbb{H}$ be a hypercomplex algebra. The product of $x$ by $y$ in $\mathbb{H}$ given by \eqref{eq:prod} satisfies the identity:
\begin{eqnarray}\label{eq5}
xy= B_{0}(x,y) + \sum_{j=1}^{n}B_{j}(x,y) \ii_j
\end{eqnarray}
where $B_{0}, B_{1},\dots , B_{n}: \mathbb{H} \times \mathbb{H}\to \mathbb{R}$ are bilinear forms whose matrix representations in the canonical base $\tau=\{1, \boldsymbol{i}_{1}, \cdots, \boldsymbol{i}_{n}\}$ are
\begin{equation}
\Big[\mathcal{B}_{0}\Big]_{\tau} = \begin{bmatrix}
1 & 0 & \cdots & 0\\
0 & p_{11,0} & \cdots & p_{1n,0}\\
\vdots & \vdots & \ddots & \vdots\\
0 & p_{n1,0} & \cdots & p_{nn,0}
\end{bmatrix} \in \mathbb{R}^{(n+1)\times(n+1)},
\end{equation}
and, for $j=1, \dots,n$,
\begin{equation}
\Big[\mathcal{B}_{j}\Big]_{\tau} = \begin{bmatrix}
0 & 0 & 0 & \cdots & 1 & \cdots & 0 &\\
0 & p_{11,j} & p_{12,j} &\cdots & p_{1j,j} & \cdots & p_{1n,j} & \\
\vdots & \vdots & \vdots & \vdots & \vdots & & \vdots &\\
1 & p_{j1,j} &p_{j2,j} & \cdots & p_{jj,j} & \cdots & p_{jn,j} & \\
\vdots & \vdots & \vdots & \vdots & \vdots & & \vdots & \\
0 & p_{n1,j} & p_{n2,j}& \cdots & p_{nj,j} & \cdots & p_{nn,j}
\end{bmatrix}  \in \mathbb{R}^{(n+1)\times(n+1)}.
\end{equation}
\end{theorem}

We note that the matrices in Theorem \ref{thm:prod_bilin} depend on the choice of basis $\tau$. Moreover, the numbers $p_{\alpha\beta,j}$ depend on the hyperimaginary unit products \eqref{eq:im_unit_prod}, which ultimately define the algebra $\mathbb{H}$.



Next we define non-degeneracy of hypercomplex algebras. From linear algebra, we have that a bilinear form is said to be non-degenerate if the following hold true
$ B(u,v)=0, \; \forall u \in \mathbb{H} \iff v=0_{ \mathbb{H}} $
and 
$ B(u,v)=0, \; \forall v \in \mathbb{H} \iff u=0_{ \mathbb{H}}. $
A bilinear form that fails this condition is degenerate. Equivalently, given the canonical basis $\tau$, a bilinear form $B$ is non-degenerate if and only if the matrix $[B]_{\tau}$ is invertible. Borrowing the terminology from linear algebra, we introduce the following definition:

\begin{definition}[Non-degenerate Hypercomplex Algebra] \label{def:non_deg}
A hypercomplex algebra $\mathbb{H}$ is non-degenerate if the matrices $[\mathcal{B}_{j}]_{\tau}$ associated with the bilinear form of the product of $\mathbb{H}$ are all invertible (see Theorem \ref{thm:prod_bilin} above). Otherwise $\mathbb{H}$ is said to be degenerate.
\end{definition}

We provide examples of Theorem \ref{thm:prod_bilin} and Definition \ref{def:non_deg} with well-known 2D hypercomplex algebras, namely the complex, hyperbolic, and dual numbers.


\begin{example}
Consider a hyperimaginary algebra $\mathbb{H}$ of dimension $2$. This algebra possesses a single hyperimaginary unit, whose product is
\begin{eqnarray*}
\boldsymbol{i}_{1}^{2}= a_{11,0} + a_{11,1}\boldsymbol{i}_{1}
\end{eqnarray*}
By computing the product of $x=x_{0} + x_{1}\boldsymbol{i}_{1}$ and $y= y_{0} + y_{1}\boldsymbol{i}_{1}$ in $\mathbb {H}$, we obtain
\begin{equation*}
xy= x_{0}y_{0} + x_{1}y_{1}a_{11,0} + (x_{1}y_{1}a_{11,1} + x_{0}y_{1} + x_{1}y_{0})\boldsymbol{i}_{1}.
\end{equation*}
Let $\tau = \{1, \boldsymbol{i}_{1}\}$ be the canonical basis of $\mathbb{H}$. From Theorem \ref{thm:prod_bilin}, the product in $\mathbb{H}$ can be written as follows
\begin{eqnarray*}
xy= \Big[x\Big]_{\tau}^{T}\Big[\mathcal{B}_{0}\Big]_{\tau}\Big[y \Big]_{\tau}  + \Big[x \Big]_{\tau}^{T}\Big[\mathcal{B}_{1}\Big]_{\tau}\Big[y \Big]_{\tau} \boldsymbol{i}_{1}.
\end{eqnarray*}
where $[x]_\tau$ and $[y]_\tau$ are the vector representation of $x$ and $y$ with respect to the canonical basis $\tau$ and the matrices of the bilinear forms are
\begin{equation*}
\Big[\mathcal{B}_{0}\Big]_{\tau} =
\begin{bmatrix}
1 & 0 \\
0 & a_{11,0}  
\end{bmatrix} \quad \text{and} \quad 
\Big[\mathcal{B}_{1}\Big]_{\tau} =
\begin{bmatrix}
0 & 1 \\
1 & a_{11,1}  
\end{bmatrix}.
\end{equation*}
In particular, we have the matrices of the bilinear forms associated with the product of complex numbers if $a_{11,0}=-1$ and $a_{11,1}=0$. Similarly, if $a_{11,0}=1$ and $ a_{11,1}=0$, we obtain the matrices of the bilinear forms associated with the product of hyperbolic numbers. Because the matrices $[\mathcal{B}_0]_\tau$ and $[\mathcal{B}_1]_\tau$ are both non-singular for either complex or hypercomplex numbers, these two algebras are notably non-degenerate. In contrast, we  have $a_{11,0} = a_{11,1} = 0$ in the product of dual numbers and, in this case, the matrix $[\mathcal{B}_0]_\tau$ is singular. Thus, the dual numbers is a degenerate hypercomplex algebra. More generally, note that $[\mathcal{B}_{1}]_{\tau}$ is non-singular regardless of the value $a_{11,1}$. Thus, the condition for a 2D hypercomplex algebra to be non-degenerate is that $[\mathcal{B}_{0}]_{\tau}$ is invertible, i.e., $a_{11,0} \neq 0$.
\end{example}

The next example addresses 4D hypercomplex algebras and include quaternions, tessarines, hyperbolic quaternions, and Klein four-group as particular instances.

\begin{example}\label{example9}
Consider a 4D hypercomplex algebra $\mathbb{H}$ in which the product of hyperimaginary units satisfies
\begin{eqnarray}
\boldsymbol{i}_{\alpha}\boldsymbol{i}_{\beta} =  a_{{\alpha \beta},0} + a_{{\alpha \beta},1}\boldsymbol{i}_{1}  + a_{{\alpha \beta},2}\boldsymbol{i}_{2} + a_{{\alpha \beta},3}\boldsymbol{i}_{3}
\end{eqnarray}
for all $\alpha, \beta \in \{1,2,3\}$. Let us take $x=x_{0} + x_{1}\boldsymbol{i}_{1} + x_{2}\boldsymbol{i}_{2} + x_{3}\boldsymbol{i}_{3} $ and $y= y_{0} + y_{1}\boldsymbol{i}_{1} + y_{2}\boldsymbol{i}_{2} + y_{3}\boldsymbol{i}_{3 } $ in $\mathbb{H}$, and the canonical basis of $\mathbb{H}$ as $\tau = \{1, \boldsymbol{i}_{1}, \boldsymbol{i}_{2}, \boldsymbol{i}_{3} \}$. Then, the product of $x$ by $y$ can be represented by bilinear forms whose matrices are given by
\begin{align*}
\Big[\mathcal{B}_{0}\Big]_{\tau} =&
\begin{bmatrix}
1 & 0 & 0 & 0\\
0 & a_{11,0} & a_{12,0}  & a_{13,0}\\ 
0 & a_{21,0} & a_{22,0}  & a_{23,0} \\
0 & a_{31,0} & a_{32,0}  & a_{33,0} 
\end{bmatrix}, 
\Big[\mathcal{B}_{1}\Big]_{\tau} =
\begin{bmatrix}
0 & 1 & 0 & 0\\
1 & a_{11,1} & a_{12,1}  & a_{13,1}\\ 
0 & a_{21,1} & a_{22,1}  & a_{23,1} \\
0 & a_{31,1} & a_{32,1}  & a_{33,1} 
\end{bmatrix},
\\ \Big[\mathcal{B}_{2}\Big]_{\tau} = &
\begin{bmatrix}
0 & 0 & 1 & 0\\
0 & a_{11,2} & a_{12,2} & a_{13,2}\\ 
1 & a_{21,2} & a_{22,2} & a_{23,2} \\
0 & a_{31,2} & a_{32,2} & a_{33,2}  
\end{bmatrix} ,
 \Big[\mathcal{B}_{3}\Big]_{\tau} = 
\begin{bmatrix}
0 & 0 &  0  & 1\\
0 & a_{11,3} &  a_{12,3}  & a_{13,3}\\ 
0 & a_{21,3} & a_{22,3} & a_{23,3} \\
1 & a_{31,3} & a_{32,3}  & a_{33,3}  
\end{bmatrix}.
\end{align*}
Therefore, an arbitrary 4D hypercomplex algebra is non-degenerate if, and only if, the above matrices are invertible. In particular the hypercomplex algebras of quaternions, tessarines, hyperbolic quaternions and Klein four-group are non-degenerate. 
\end{example}

\subsection{Universal Approximation Theorem to a Broad Class of Hypercomplex-valued Neural Networks}


In the previous sections we have presented a few universal approximation theorems. A common theme among them is the requirement for the activation function to be discriminatory. We have also defined degeneracy of hypercomplex algebras. The main result of this work, namely, the Universal Approximation Theorem for a broad class of hypercomplex-valued neural networks is achieved by combining these concepts and properties. In this section we formalize a few definitions and notations before stating our result in Theorem \ref{thm:tau_geral}.

We start off by recalling that a split activation function $\psi_{\mathbb{H}} : \mathbb{H} \rightarrow \mathbb{H}$ is defined based on a real function $\psi : \mathbb{R} \rightarrow \mathbb{R}$ by
\begin{equation}\label{eq:split_act_f}
\psi_{\mathbb{H}}(x) = \psi(x_0) + \boldsymbol{i}_{1} \psi(x_{1}) + \boldsymbol{i}_{2} \psi(x_{1}) + \dots + \boldsymbol{i}_{n} \psi(x_{2})
\end{equation}
for all $x=x_{0} + x_{1}\boldsymbol{i}_{1} + \cdots x_{n}\boldsymbol{i}_{n} \in \mathbb{H}$. In this work, the activation functions chosen are the split $\mathtt{ReLU}$ and the split $\mathtt{sigmoid}$, both well-known from applications and from the literature of other approximation theorems.

We define an $\mathbb{H}$MLP as a MLP model in which inputs, outputs, and trainable parameters are hypercomplex numbers instead of real numbers. By making such a general definition we encompass previously known models such as complex, quaternion, hyperbolic, tessarine, and Clifford-valued networks as particular cases, thus resulting in a broader family of models. In the following definition we highlight that in hypercomplex-valued MLPs the feedforward step can also be seen as a finite linear combination.



\begin{definition}[$\mathbb{H}$MLP]\label{def:HMLP}
Let $\mathbb{H}$ be a hypercomplex algebra. A hypercomplex-valued multilayer perceptron ($\mathbb{H}$MLP) can be described by 
\begin{equation}\label{eq:hmlp}
\mathcal{N}_\mathbb{H}(\boldsymbol{x}) = \sum_{i=1}^{M} \alpha_{i} \psi(\boldsymbol{y}_{i}^{T} \cdot \boldsymbol{x} + \theta_{i}), \forall \boldsymbol{x} \in \mathbb{H}^N ,
\end{equation}
where $\boldsymbol{x} \in \mathbb{H}^{N}$ represents the input to the neural network, $\mathcal{N}_\mathbb{R}(\boldsymbol{x}) \in \mathbb{H}$ is the output, $y_{i} \in \mathbb{H}^{N}$ and $\alpha_{i} \in \mathbb{H}$ are the weights between input and hidden layers, and hidden and output layers, respectively, $\theta_{i} \in \mathbb{H}$ are the biases for the neurons in the hidden layer, and $\psi: \mathbb{H} \to \mathbb{H}$ is the activation function. The number of neurons in the hidden layer is $M$.
\end{definition}
This definition is analogous to the real-valued MLP described in Section \ref{sec:UATS_diversos}. 
Now, we have the necessary components and can state the main result of this work: the extension of the universal approximation theorem to neural networks defined in non-degenerate  hypercomplex algebras.

\begin{theorem}\label{thm:tau_geral}
Consider a non-degenerate hypercomplex algebra $\mathbb{H}$ and let $K \subset \mathbb{H}^N$ be a compact. Also, consider a real-valued continuous discriminatory function $\psi: \mathbb{R} \to \mathbb{R}$ such that $\lim_{\lambda  \to -\infty} \psi(\lambda)=0$ and let $\psi_{\mathbb{H}}: \mathbb{H} \to \mathbb{H}$ be the split function associated to $\psi$ by means of \eqref{eq:split_act_f}. Then, the class 
\begin{equation}\label{eq:class-HMLP}
    \mathcal{H}_\mathbb{\psi} = \left\{ \mathcal{N}_\mathbb{H}(\boldsymbol{x}) = \sum_{i=1}^{M} \alpha_{i} \psi(\boldsymbol{y}_{i}^{T} \cdot \boldsymbol{x} + \theta_{i}): M \in \mathbb{N}, \boldsymbol{y}_i \in \mathbb{H}^N, \alpha_i, \theta_i \in \mathbb{H} \right\},
\end{equation}
is dense in the set $\mathcal{C}(K)$ of all hypercomplex-valued continuous functions on $K$. In other words, given a hypercomplex-valued continuous function $f_\mathbb{H}:K \to \mathbb{H}$ and $\epsilon>0$, there exists a $\mathbb{H}$MLP network $\mathcal{N}_\mathbb{H}:\mathbb{H}^N \to \mathbb{H}$ given by \eqref{eq:hmlp} such that
\begin{equation}
    | f_\mathbb{H}(\boldsymbol{x})-\mathcal{N}_\mathbb{H}(\boldsymbol{x})|< \epsilon, \quad \forall \boldsymbol{x} \in K,
\end{equation}
where $|\cdot|$ denotes the absolute value of hypercomplex numbers defined by \eqref{eq:absolute_value}.
\end{theorem}

The Theorem \ref{thm:tau_geral} extends the existing universal approximation theorems \cite{arena1998neural,buchholz2000hyperbolic,Buchholz2001,eniac,ARENA1997335} and strengthens neural networks models on the broad family of non-degenerate hypercomplex algebras.


\section{Concluding Remarks}\label{sec:concluding}

The universal approximation theorem asserts that a single hidden layer neural network can approximate continuous functions with arbitrary precision. This essential theoretical result was first proven for real-valued networks in the late 1980s \cite{Cybenko1989ApproximationFunction}. In the years that followed, the universal approximation theorem was also proven for neural networks based on well-known hypercomplex algebras, such as complex \cite{arena1998neural}, quaternions \cite{ARENA1997335}, and Clifford algebras \cite{Buchholz2001}. However, each of these results was derived individually, meaning there is a lack of generality in the proofs of universal approximation theorems. In this work, we investigate the existing theorems and tie the universal approximation property of hypercomplex-valued networks to two main factors: an appropriate activation function choice and the underlying algebra's degeneracy. By identifying these objects, we review the definitions of discriminatory activation functions and introduce the concept of non-degenerate hypercomplex algebras. Finally, we give sufficient conditions for a neural network to be a universal approximator in a broad class of hypercomplex-valued algebras. Specifically, we formulate the universal approximation theorem: hypercomplex-valued single hidden layer neural networks with discriminatory split activation functions are dense in the set of continuous functions on a compact subset of the Cartesian product of a non-degenerate hypercomplex algebra. 

The universal approximation theorem formulated in this paper serves many purposes, including the following items:
\begin{enumerate}
    \item It consolidates the results regarding the universal approximation property of many well-known algebras, thus eliminating the need to prove this property for each algebra individually. In particular, the class of non-degenerate hypercomplex algebras includes the complex and hyperbolic numbers, quaternions, tessarines, and Clifford algebras, all of which have particular results of their own, as mentioned in previous sections.
    \item Many algebras that have not had this result proven are now directly known as the basis for neural networks with universal approximation property. That is the case for the Klein group and the octonions, among others.
    \item This result further promotes the use of hypercomplex-valued networks. Indeed, hypercomplex-valued networks are known to perform well in problems involving multidimensional signals such as images, video, and 3D movement \cite{parcollet2020survey,vieira2022general}. The property of universal approximators strengthens these models' applications, posing them as strictly better than real-valued models for a wider variety of applications.
\end{enumerate}


\begin{thebibliography}{10}
\providecommand{\url}[1]{\texttt{#1}}
\providecommand{\urlprefix}{URL }
\providecommand{\doi}[1]{https://doi.org/#1}

\bibitem{albert42}
Albert, A.A.: {Quadratic Forms Permitting Composition}. Annals of Mathematics
  \textbf{43}(1),  161--177 (1942)

\bibitem{alfsmann2006families}
Alfsmann, D.: On families of 2 n-dimensional hypercomplex algebras suitable for
  digital signal processing. In: 2006 14th European Signal Processing
  Conference. pp.~1--4. IEEE (2006)

\bibitem{ARENA1997335}
Arena, P., Fortuna, L., Muscato, G., Xibilia, M.: Multilayer perceptrons to
  approximate quaternion valued functions. Neural Networks  \textbf{10}(2),
  335--342 (1 1997)

\bibitem{arena1998neural}
Arena, P., Fortuna, L., Muscato, G., Xibilia, M.G.: Neural networks in
  multidimensional domains: fundamentals and new trends in modeling and
  control. Springer London (1998)

\bibitem{Breuils2022NewAlgebra}
Breuils, S., Tachibana, K., Hitzer, E.: {New Applications of Clifford’s
  Geometric Algebra}. Advances in Applied Clifford Algebras 2022 32:2
  \textbf{32}(2),  1--39 (2 2022). \doi{10.1007/S00006-021-01196-7},
  \url{https://link.springer.com/article/10.1007/s00006-021-01196-7}

\bibitem{buchholz2000hyperbolic}
Buchholz, S., Sommer, G.: A hyperbolic multilayer perceptron. In: Proceedings
  of the IEEE-INNS-ENNS International Joint Conference on Neural Networks.
  IJCNN 2000. Neural Computing: New Challenges and Perspectives for the New
  Millennium. vol.~2, pp. 129--133. IEEE (jul 2000)

\bibitem{Buchholz2001}
Buchholz, S., Sommer, G.: Clifford Algebra Multilayer Perceptrons, pp.
  315--334. Springer Berlin Heidelberg, Berlin, Heidelberg (2001)

\bibitem{Buchholz2008OnPerceptrons}
Buchholz, S., Sommer, G.: {On Clifford neurons and Clifford multi-layer
  perceptrons}. Neural Networks  \textbf{21}(7),  925--935 (9 2008).
  \doi{10.1016/j.neunet.2008.03.004}

\bibitem{eniac}
Carniello, R., Vital, W., Valle, M.: Universal approximation theorem for
  tessarine-valued neural networks. In: Anais do XVIII Encontro Nacional de
  Inteligência Artificial e Computacional. pp. 233--243. SBC, Porto Alegre,
  RS, Brasil (2021). \doi{10.5753/eniac.2021.18256},
  \url{https://sol.sbc.org.br/index.php/eniac/article/view/18256}

\bibitem{Catoni2008TheSpace-Time}
Catoni, F., Boccaletti, D., Cannata, R., Catoni, V., Nichelatti, E., Zampetti,
  P.: {The Mathematics of Minkowski Space-Time}. Birkh{\"{a}}user Basel (2008).
  \doi{10.1007/978-3-7643-8614-6}

\bibitem{CERRONI2017232}
Cerroni, C.: From the theory of congeneric surd equations to segre's bicomplex
  numbers. Historia Mathematica  \textbf{44}(3),  232--251 (2017).
  \doi{https://doi.org/10.1016/j.hm.2017.03.001},
  \url{https://www.sciencedirect.com/science/article/pii/S0315086017300241}

\bibitem{Cybenko1989ApproximationFunction}
Cybenko, G.: {Approximation by superpositions of a sigmoidal function}.
  Mathematics of Control, Signals and Systems 1989 2:4  \textbf{2}(4),
  303--314 (12 1989),
  \url{https://link.springer.com/article/10.1007/BF02551274}

\bibitem{Arnold2009}
Givental, A.B., Khesin, B.A., Marsden, J.E., Varchenko, A.N., Vassiliev, V.A.,
  Viro, O.Y., Zakalyukin, V.M. (eds.): On functions of three variables,
  pp.~5--8. Springer Berlin Heidelberg, Berlin, Heidelberg (2009),
  \url{https://doi.org/10.1007/978-3-642-01742-1_2}

\bibitem{grassucci22prl}
Grassucci, E., Mancini, G., Brignone, C., Uncini, A., Comminiello, D.: Dual
  quaternion ambisonics array for six-degree-of-freedom acoustic representation
  (2022). \doi{10.48550/ARXIV.2204.01851}

\bibitem{grassucci2022lightweight}
Grassucci, E., Zhang, A., Comminiello, D.: Lightweight convolutional neural
  networks by hypercomplex parameterization (2022),
  \url{https://openreview.net/forum?id=S5qdnMhf7R}

\bibitem{guilhoto2018overview}
Guilhoto, L.F.: An overview of artificial neural networks for mathematicians
  (2018)

\bibitem{hitzer13}
Hitzer, E., Nitta, T., Kuroe, Y.: {Applications of Clifford's Geometric
  Algebra}. Advances in Applied Clifford Algebras  \textbf{23}(2),  377--404 (6
  2013). \doi{10.1007/s00006-013-0378-4}

\bibitem{hoffman1971linear}
Hoffman, K.: Linear algebra. Englewood Cliffs, NJ, Prentice-Hall (1971)

\bibitem{HORNIK1991251}
Hornik, K.: Approximation capabilities of multilayer feedforward networks.
  Neural Networks  \textbf{4}(2),  251--257 (1991).
  \doi{https://doi.org/10.1016/0893-6080(91)90009-T},
  \url{https://www.sciencedirect.com/science/article/pii/089360809190009T}

\bibitem{huang2013klein}
Huang, J.S., Yu, J.: Klein four-subgroups of lie algebra automorphisms. Pacific
  Journal of Mathematics  \textbf{262}(2),  397--420 (2013)

\bibitem{kantor1989hypercomplex}
Kantor, I., Solodovnikov, A.: Hypercomplex numbers: an elementary introduction
  to algebras, vol.~302. Vol. 302. New York: Springer-Verlag, (1989)

\bibitem{KOBAYASHI2020123}
Kobayashi, M.: Hopfield neural networks using klein four-group. Neurocomputing
  \textbf{387},  123--128 (2020).
  \doi{https://doi.org/10.1016/j.neucom.2019.12.127},
  \url{https://www.sciencedirect.com/science/article/pii/S0925231220300850}

\bibitem{kolmogorov1957representation}
Kolmogorov, A.N.: On the representation of continuous functions of many
  variables by superposition of continuous functions of one variable and
  addition. In: Doklady Akademii Nauk. vol.~114, pp. 953--956. Russian Academy
  of Sciences (1957)

\bibitem{McCulloch1943}
McCulloch, W.S., Pitts, W.: {A logical calculus of the ideas immanent in
  nervous activity}. The bulletin of mathematical biophysics  \textbf{5}(4),
  115--133 (12 1943),
  \url{https://link.springer.com/article/10.1007/BF02478259}

\bibitem{parcollet2020survey}
Parcollet, T., Morchid, M., Linar{\`e}s, G.: A survey of quaternion neural
  networks. Artificial Intelligence Review  \textbf{53}(4),  2957--2982 (4
  2020)

\bibitem{1306653}
Pei, S.C., Chang, J.H., Ding, J.J.: Commutative reduced biquaternions and their
  fourier transform for signal and image processing applications. IEEE
  Transactions on Signal Processing  \textbf{52}(7),  2012--2031 (2004).
  \doi{10.1109/TSP.2004.828901}

\bibitem{PETERSEN2018296}
Petersen, P., Voigtlaender, F.: Optimal approximation of piecewise smooth
  functions using deep relu neural networks. Neural Networks  \textbf{108},
  296--330 (2018). \doi{https://doi.org/10.1016/j.neunet.2018.08.019},
  \url{https://www.sciencedirect.com/science/article/pii/S0893608018302454}

\bibitem{rosenblatt1958perceptron}
Rosenblatt, F.: {The perceptron: a probabilistic model for information storage
  and organization in the brain.} Psychological review  \textbf{65}(6),
  386--408 (12 1958),
  \url{https://link.springer.com/article/10.1007/BF02551274}

\bibitem{Eniacc}
Senna, F., Valle, M.: Tessarine and quaternion-valued deep neural networks for
  image classification. In: Anais do XVIII Encontro Nacional de Inteligência
  Artificial e Computacional. pp. 350--361. SBC, Porto Alegre, RS, Brasil
  (2021). \doi{10.5753/eniac.2021.18266},
  \url{https://sol.sbc.org.br/index.php/eniac/article/view/18266}

\bibitem{Takahashi2021ComparisonControl}
Takahashi, K.: {Comparison of high-dimensional neural networks using
  hypercomplex numbers in a robot manipulator control}. Artificial Life and
  Robotics  \textbf{26}(3),  367--377 (8 2021)

\bibitem{vaz16}
Vaz, J., da~Rocha, R.: {An Introduction to Clifford Algebras and Spinors}.
  Oxford University Press (2016)

\bibitem{vieira22wcci_arxiv}
Vieira, G., Valle, M.E.: Acute lymphoblastic leukemia detection using
  hypercomplex-valued convolutional neural networks (2022).
  \doi{10.48550/ARXIV.2205.13273}

\bibitem{vieira2022general}
Vieira, G., Valle, M.E.: A general framework for hypercomplex-valued extreme
  learning machines. Journal of Computational Mathematics and Data Science
  \textbf{3},  100032 (2022)

\end{thebibliography}
\end{document}